# Direct Uncertainty Estimation in Reinforcement Learning

Sergey Rodionov[1,2], Alexey Potapov[1,3], Yurii Vinogradov[3]

[1]AIDEUS, Russia
[2]Aix Marseille Université, CNRS, LAM (Laboratoire d'Astrophysique de Marseille) UMR 7326, 13388, Marseille, France
[3]National Research University of Information Technology, Mechanics and Optics, St. Petersburg, Russia
{potapov,rodionov}@aideus.com

Optimal probabilistic approach in reinforcement learning is computationally infeasible. Its simplification consisting in neglecting difference between true environment and its model estimated using limited number of observations causes exploration vs exploitation problem. Uncertainty can be expressed in terms of a probability distribution over the space of environment models, and this uncertainty can be propagated to the action-value function via Bellman iterations, which are computationally insufficiently efficient though. We consider possibility of directly measuring uncertainty of the action-value function, and analyze sufficiency of this facilitated approach.

## 1  Introduction

Absence of prior knowledge about the environment (absence of its precise model) is naturally characterized by the intuitive notion of uncertainty. However, no generally accepted accurate formal description of this notion exists. Probability theory is the most traditional way of describing uncertainty, but adequate interpretation of probability itself is not that clear. This can be seen from numerous paradoxes in probability theory, such as the grue emerald paradox. That is why some attempts of extending probability theory were made. The most well-known one is fuzzy set theory. However, fuzzy operations can be considered as probabilistic operations with some additional assumptions about operands (e.g. their independence), which allow simplifying their computation. The main difference here lies not in formalisms, but in their interpretation and usage.

Differences in interpretations frequently appear, because complex systems (e.g. verbal notions) are being analyzed. Some simple measure of uncertainty cannot be applied in such cases without being a part of some model of intelligence. Thus, consideration of this problem in the context of intelligent agents can be most appropriate. These agents can be put in environments of different classes, e.g. Markov environments or arbitrary computable environments. The latter are quite interesting since they possess maximum uncertainty. Exactly this case helps to reveal difficulties in classic probability theory (e.g. with assignment of prior probabilities or probabilities of unique events), which are solved within algorithmic probability theory [1]. Optimal solutions with similar structure exist both for Markov environments using classic probability and for computable environments using algorithmic probability without involving other formalisms.

However, this solution is computationally intractable, since it consists in enumeration of all possible actions and all possible responses of the environment with estimation of probability of each possibility. Traditional reinforcement learning methods rely only on the most probable model of the environment ignoring all other models, which makes them computationally very efficient. The same simplification can be applied also to such models as AIXI [2] dealing with arbitrary environments, but it is insufficient for achieving efficient universal agents, so it has not been analyzed in detail. We also consider this simplification on the example of Markov environments assuming that the case of arbitrary environments mainly differs in the way of computing probabilities. Policies that use only one best model of the environment for prediction lack exploration in classic RL [3]. This is intuitively clear since the agent acting on the base of the single model implicitly assumes that it knows everything about the environment. This lack is compensated not by using many models, but by introducing specific exploratory strategies (simplest one is ε-greedy). However, this results in the exploration vs exploitation dilemma. This dilemma is exactly the reason for introducing different measures of uncertainty in RL methods. Hence uncertainty is introduced as a technique for compensating negative effects caused by simplification of optimal probabilistic methods. More specifically, the agent can use only one model, but accounting for its uncertainty, and can perform exploratory actions in order to reduce this uncertainty.

In this paper, we consider one specific approach of introducing and utilizing uncertainty in models of RL agents in Markov environments. This approach relies on direct empirical estimation of uncertainty for a "split" action-value function $Q(s, a, s')$. It shows adequate results in solving the exploration vs exploitation dilemma meaning that uncertainty can indeed be considered as a "heuristic" that helps to greatly simplify the optimal solution without violating its important features.

## 2 Related Works

Consider traditional settings for RL-agents [3]. Let an RL-agent be placed in a Markov environment defined by a state space $S$, a set of possible actions $A$, transition probabilities $P(s'|s, a)$, $P: S \times A \times S \to [0,1]$, and a reward function $R(s, a, s')$, $R: S \times A \times S \to \mathbf{R}$, where $\mathbf{R}$ is a set of possible reward values.

The value function $V^\pi(s)$ for some policy $\pi$ is calculated as the summed discounted future rewards

$$V^\pi(s) = E_t\left(\sum_t \gamma^t R(s_t, \pi(s_t), s_{t+1})\right),$$

and it should be maximized over some policy space. Here, γ is the conventional discount factor [3].

The action-value function $Q^\pi(s, a)$ that depends also on the action $a$ is usually introduced. The action-value function $Q^*$ for optimal strategy should satisfy the well-known Bellman equation

$$Q^*(s,a) = E_{s'}\left(R(s,a,s') + \gamma \max_{a'} Q^*(s',a')\right).$$

If the true model of the environment including $P$ and $R$ is known, $Q^*$ can be estimated by dynamic programming. If such model is not given, it seems natural to use some estimation of $P$ and $R$, and this is done in most existing RL-methods (estimated $P$ and $R$ are not necessarily used in the explicit form). Transition probabilities are simply replaced by frequencies. However, different values of transition probabilities are also possible. Indeed, if some action was used $n$ times in some state, and it had $i$-th outcome $n_i$ times, the probability $P(p_i | n_i, n)$ that real probability of $i$-th outcome given these observations equals $p_i$ can be calculated using Bayes' rule

$$P(p_i | n_i, n) \propto P(n_i, n | p_i) P(p_i) = p_i^{n_i} (1-p_i)^{n-n_i} P(p_i), \qquad (1)$$

where $P(p_i)$ are priors (without prior information one can assume that $P(p_i)=P(p_j)$ for any $p_i=p_j$ taking into account that $\sum_i p_i = 1$).

Bayesian agents should account for these probabilities of probabilities and their updates after new hypothetical observations in order to take optimal actions. If these probabilities are ignored, corresponding agent will diverge from the optimal behavior. However, their maintenance is computationally infeasible, and some simplifications are needed. For example, uncertainty in transition probabilities is represented by covariance matrices in [4]. That is, each transition probability is treated as a normally distributed random variable. Uncertainty in $R$ is also taken into account (this uncertainty arises for $s$-$a$-$s'$ transitions that have never taken place yet). The Bellman equation written for random variables is used to propagate uncertainty from $P$ and $R$ to $Q$ using dynamic programming.

In [5], this approach is simplified by ignoring covariances (only dispersions are used). However, this method still requires propagation of uncertainty using Bellman iterations on each step. This is not too practical. One would like to update uncertainty of $Q$ as easy as updates of $Q$ are done in such methods as Q-learning or SARSA.

Another question is connected to adoption of uncertainty in selecting actions. Authors of [4] proposed to introduce some modified action-value function for uncertainty-aware policy improvement

$$Q'(s,a) = Q(s,a) - \xi \sigma Q(s,a), \qquad (2)$$

where ξ is the parameter for balancing exploration and exploitation, and $\sigma Q(s, a)$ is the standard deviation of the random variable $Q(s, a)$ described by normal distribution. Presence of the free parameter ξ indicates that this is not a complete solution of the exploration vs exploitation dilemma.

Other approaches to incorporating uncertainty in RL exist (see references in [4, 5]), but the approach mentioned above is the most appropriate for our objectives.

## 3  Direct Estimation of Uncertainty of $Q$

Uncertainty propagation has a clear sense, but it removes the main advantage of model-free reinforcement learning. This advantage consists in the fact that $Q(s, a)$ has lower dimension than $P(s, a, s')$ and thus it should require less data to be learned. Is it possible to estimate uncertainty of $Q$ without explicitly constructing environment models as it can be done while estimating $Q$ itself? Or at least is it possible to avoid uncertainty propagation?

One can try evaluating variations of $Q(s, a)$ for each pair $(s, a)$ empirically. Indeed, changes of $Q(s, a)$ can be caused by the lack of agent's knowledge. Thus, the agent should simply accumulate averaged $Q_t^2(s, a)$ in addition to averaged $Q_t(s, a)$ for each $(s, a)$ to calculate dispersion of $Q_t(s, a)$ indicating its uncertainty. Unfortunately, this approach works only in deterministic environments. The reason is quite clear. Stochasticity of the environment causes persistent variations of $Q$ values. Consider the following well-known update rule as an example

$$Q(s_t, a_t) \leftarrow Q(s_t, a_t) + \alpha [r_t + \gamma \max_a Q(s_{t+1}, a) - Q(s_t, a_t)], \qquad (3)$$

where α is the learning rate.

Even if optimal values $Q^*(s, a)$ are used as initial values, updates will cause variations of the deviation $\sigma Q(s, a) \propto \alpha \sigma_{s'} R(s' | s, a)$.

Thus, variations of $Q(s, a)$ caused by imprecise knowledge of the environment and by intrinsic stochasticity of the environment should be separated (these two types of uncertainty of $Q$ have entirely different meaning, but both of them can be expressed in terms of probability).

To do this, we introduce "split" action-value functions $Q(s, a, s')$, which variations are caused only by uncertainty of the agent's knowledge, but not stochasticity of the environment. Indeed, this function can be updated in the same way as $Q(s, a)$

$$Q(s_t, a_t, s_{t+1}) \leftarrow Q(s_t, a_t, s_{t+1}) + \alpha [r_t + \gamma \max_a Q(s_{t+1}, a) - Q(s_t, a_t, s_{t+1})]. \qquad (4)$$

Since $r_t$ will always be the same for certain $s$-$a$-$s'$ transition, stochasticity of the environment will not cause variations of $Q(s, a, s')$. Unfortunately, the total action-value function $Q(s, a)$ should be calculated as

$$Q(s, a) = \sum_{s'} P(s' | s, a) Q(s, a, s'). \qquad (5)$$

Thus, transition probabilities should be estimated in any case, but the uncertainty $\sigma Q(s, a, s')$ can be estimated empirically without its propagation via dynamic programming.

Now let us consider a more sensible way of adopting uncertainty in selecting action than (2). If it is assumed that *Q*-values are random variables, for which Bellman equation can be written, than it should be also assumed that such equations for selecting actions as

$$\pi(s) = \arg\max_a Q(s,a)$$

are written for random variables making π(*s*) also a random variable.

Since *Q*(*s*, *a*) has some estimated probability distribution (e.g. Gaussian distribution described by *EQ*(*s*, *a*) and σ*Q*(*s*, *a*)), one can simply generate one sample of *Q*(*s*, *a*) for each *a* in accordance with corresponding distributions and find maximum among them to choose an action. This procedure will output realizations of the random variable $\arg\max_a Q(s,a)$. Thus, it is the correct extension of the reward-maximization policy on the case of uncertain *Q*.

However, uncertainty is estimated for *Q*(*s*, *a*, *s'*) in our case, and uncertainty adoption should be slightly more complicated. Random values of *Q*(*s*, *a*, *s'*) for each *s'* are sampled from corresponding distributions and their weighted average value is calculated as an instance of *Q*(*s*, *a*). Weights are uncertain probabilities in (5), which specific values are also randomly sampled in accordance with the distribution (1). The acceptance-rejection method is used here: $p_i$ are randomly sampled from [0, 1]; they are normalized (their sum should equal 1.0); then they are accepted if the value randomly chosen from the range [0, 1] is smaller than their likelihood (prior probabilities are assumed to be uniformly distributed).

The vaguest issue is accounting for transitions that have not been encountered yet. One should estimate possible probabilities of such transitions and their Q-values. Our agent assumes that there is one unknown outcome for each action. Its probability can easily be estimated using (1) as $P(p_{unknown} \mid 0, n) \propto (1 - p_{unknown})^n$. However, there is no information about *Q*(*s*, *a*, *s'*) and its uncertainty for this hypothetical transition. This uncertainty is "absolute", i.e. there is no universal prior distribution of *Q* for arbitrary environment. One can consider the lottery environment, in which the agent has two actions corresponding to buying a ticket or not. Both probability of winning and possible prize are unknown. One can imagine two types of the environment with positive and negative expected reward for the first action (the second action has zero reward). Imagine that the agent has bought losing tickets for $10^{100}$ times. Should it stop? It is impossible to answer this question without any prior knowledge about the distribution of *Q* (or *R*). We should incorporate such prior knowledge about the real world into AGI (artificial general intelligence). However, in this paper, we simply applied optimistic approach [3] assuming that the largest possible reward is known. We also initialize Q-values with maximum value of *Q* in Q-learning methods that greatly encourages exploration even with small ε.

## 4    Experiments

In our experiments, we consider "layered" environments with the following structure. Level zero (*l*=0) has one state; all other levels (*l*=1 ...*m*) have *n* states per level. Single state on level zero (*l*=0), has *n* possible actions, and each of them leads with probability *p*=1.0 to corresponding state on *l*=1. Each state on the last level (*l*=*m*) has only one possible action, which leads to the single state on level zero with *p*=1.0. Each state on intermediate levels *l*=1...*m*–1 has *k* possible actions, each of which has two possible results leading to one of two states on *l*=*i*+1 (probabilities of possible outcomes of each action are chosen randomly). Here we used *m*=20, *n*=10, *k*=2.

We compared three algorithms: (1) ordinary Q-learning with ε-greedy strategy, which is turned off after some large number of steps (when Q-function is learned); (2) modified Q-learning (we will refer to it as split-Q-learning) with the split action-value function *Q*(*s*, *a*, *s'*) and also with disabling ε-greedy strategy; (3) modified Q-learning with uncertain *Q*(*s*, *a*, *s'*).

It should be pointed out that disabling ε-greedy strategy is "unfair", since we do not know for arbitrary environment, when exploration should be stopped (and in the case of non-stationary environments it should not be stopped at all). Thus, rewards during exploration show performance of all methods, and rewards after terminating ε-greedy strategy of the first two methods serve as the reference result for the third algorithm that continues exploration, because uncertainty is not reduced to zero.

Fig. 1 shows rewards (averaged over 10000 trial environments) obtained by these three algorithms on each step.

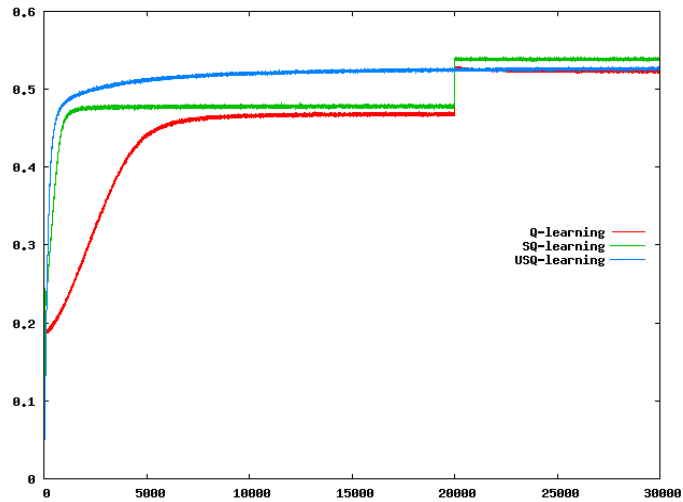

**Fig. 1.** Average rewards for ordinary Q-learning, split-Q-learning, uncertain split-Q-learning

It can be seen that split-Q-learning outperforms ordinary Q-learning. This result deserves independent study that goes beyond the topic of this paper. Discontinuity on the curves obtained for two first algorithms is caused by turning off the ε-greedy strategy. Interestingly, rewards of ordinary Q-learning after this step are lower than that of split-Q-learning. The possible reason consists in constant updates of Q-values using the latest rewards, which are random in stochastic environments (and thus Q-function oscillates). Indeed, this difference appears to be smaller for lower values of α parameter, but smaller α also causes slower convergence.

Uncertain split-Q-learning shows better performance in comparison with the ε-greedy strategy. One can try different values of ε. Higher values should yield faster exploration and achievement of the optimal policy, but this policy will be "penalized" by large number of random action (until turning off the ε-greedy strategy). Smaller values of ε will converge slower (but quite fast in the case of optimistic agents) to higher level of rewards. In our experiments, uncertainty-based exploration outperformed the ε-greedy strategy with both small and large values of ε meaning that it goes faster to higher level of rewards. This is equivalent to adaptive alteration of ε from initial high values to subsequent small values. Of course, it does not reach the maximum possible level of rewards (in comparison with the optimal policy with the turned off ε-greedy strategy), because some uncertainty always remains.

## Conclusion

The approach for direct empirical estimation of uncertainty in models of RL agents was introduced on the base of the split action-value function $Q(s, a, s')$. This approach does not require uncertainty propagation and can be computationally efficient. Preliminary experiments with this approach showed possibility of solving the exploration vs exploitation problem without considering full probability distribution over the space of all environment models.

The main issue for future investigations is connected with non-stationary environments, for which uncertainty should not converge to zero and should supply an appropriate level of explorations.

## Acknowledgements

This work was supported by the Ministry of Education and Science of the Russian Federation.